\PassOptionsToPackage{table}{xcolor}
\RequirePackage{fix-cm}
\documentclass{amap}
\usepackage[utf8]{inputenc}
\usepackage[T1]{fontenc}
\usepackage{booktabs,multirow,bm,colortbl}
\usepackage{amsmath,amssymb,amsfonts}
\usepackage{graphicx,xcolor,microtype,xspace}
\usepackage{hyperref,url}
\newcommand{\methodname}{\text{ActionPiece}\xspace}

\newcommand{\liberoOurs}{94.8}
\newcommand{\liberoBestBaseline}{93.7}
\newcommand{\plusOurs}{68.8}
\newcommand{\plusBestBaseline}{64.3}

\newcommand{\vlaLZero}{82.2}
\newcommand{\vlaLOne}{42.7}
\newcommand{\vlaLTwo}{29.5}
\newcommand{\vlaMean}{51.5}

\newcommand{\simplerOurs}{71.9}

\newcommand{\papertitle}{ActionPiece: Rethinking Action Tokenization for Autoregressive Vision-Language-Action Models}

\usepackage{adjustbox,placeins}
\newcommand{\papertablestyle}{%
  \small
  \renewcommand{\textbf}[1]{{\bfseries ##1}}%
  \renewcommand{\arraystretch}{1.18}%
  \setlength{\tabcolsep}{4pt}%
}

\usepackage[bottom]{footmisc}
\patchcmd{\mymaketitle}{\vskip 7mm}{\vskip 0mm}{}%
  {\PackageError{company-layout}{Could not adjust title spacing}{Check amap.cls.}}

\patchcmd{\mymaketitle}{\abstractlist\par}{\abstractlist\par\vspace{-3mm}}{}%
  {\PackageError{company-layout}{Could not adjust abstract link spacing}{Check amap.cls.}}
\patchcmd{\mymaketitle}{boxrule=0.5pt,arc=2.5mm}{boxrule=0.5pt,arc=2.5mm,bottom=2mm}{}%
  {\PackageError{company-layout}{Could not adjust abstract box padding}{Check amap.cls.}}
\fancypagestyle{firststyle}{%
  \fancyhf{}%
  \fancyhead[L]{\vskip 2mm\includegraphics[height=7.5mm]{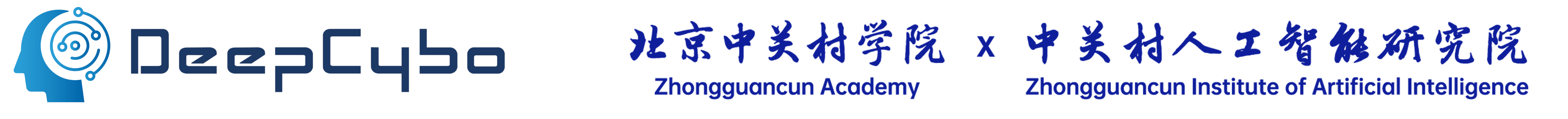}}%
  \fancyfoot[C]{\thepage}%
}

\renewcommand{\authorfont}{\fontfamily{ptm}\fontseries{m}\fontsize{11}{13}\selectfont}
\renewcommand{\authorformat}[2][]{\mbox{\authorfont #2\textsuperscript{\ensuremath{#1}}}}
\renewcommand{\affiliationfont}{\fontfamily{ptm}\fontseries{m}\fontsize{9}{11}\selectfont}
\renewcommand{\affiliationformat}[2][]{{\affiliationfont\textsuperscript{\ensuremath{#1}}#2}}
\patchcmd{\mymaketitle}{\vskip 1.2em}{\vskip 0.7em}{}%
  {\PackageError{company-layout}{Could not adjust author spacing}{Check amap.cls.}}

\title{\papertitle}
\author[1,2,3]{Shijie Lian}
\author[4,2,3]{Bin Yu}
\author[5,2,3]{Zhaolong Shen}
\author[6,3]{Xiaopeng Lin}
\author[3]{Yichao Du}
\author[3,\ddagger]{Zhirui Zhang}
\author[1,7,\dagger]{Laurence T. Yang}
\author[8,3,\dagger]{Kai Chen}
\affiliation[1]{Huazhong University of Science and Technology}
\affiliation[2]{Zhongguancun Academy}
\affiliation[3]{DeepCybo}
\affiliation[4]{Harbin Institute of Technology}
\affiliation[5]{Beihang University}
\affiliation[6]{The Hong Kong University of Science and Technology (Guangzhou)}
\affiliation[7]{Zhengzhou University}
\affiliation[8]{Zhongguancun Institute of Artificial Intelligence}

\newcommand{\companyauthornotes}{%
  \begingroup
  \renewcommand{\thefootnote}{\fnsymbol{footnote}}%
  \footnotetext[3]{Project Leader.}%
  \footnotetext[2]{Corresponding authors.}%
  \renewcommand{\thefootnote}{}%
  \footnotetext{Work done during internship at DeepCybo.}%
  \endgroup
}

\projectpage{https://deepcybo-physai.github.io/ActionPiece/}
\abstract{Action tokenizers play a central role in autoregressive vision-language-action (VLA) models, determining both the targets for policy training and the executable commands recovered from predicted tokens. Their fidelity is commonly evaluated using pointwise reconstruction metrics such as mean squared error (MSE), yet small individual errors do not fully characterize how faithfully action adjustments across demonstrations are preserved. After compression, similar actions may still cluster around a representative motion, while the adjustments needed for different contexts are diminished, distorted, or even reversed. We introduce physical rank consistency (PRC) to measure how well tokenization preserves local physical distance rankings after reconstruction. Evaluating decoded actions provides a common reference across token vocabularies and decoder architectures, complementing pointwise accuracy with a measure of relational fidelity. We further present \methodname, which preserves physical action relationships through joint supervision of representation learning and quantization. Physical rank preservation supervises near--far ordering in encoder and quantized feature distances, while quantization regularization applies the same ordering to codeword assignment distributions. Both objectives augment reconstruction, producing discrete action tokens for standard autoregressive policy learning and execution through a frozen decoder. Under the same Qwen3-VL-4B policy training setup, \methodname\ achieves \liberoOurs\% on LIBERO and \plusOurs\% on unseen LIBERO-Plus, with additional evaluations reaching \simplerOurs\% on SimplerEnv and \vlaMean\% across VLA-Arena L0--L2. Component ablations show that the two objectives jointly improve PRC and policy success, demonstrating the value of physical relationship supervision for action tokenization.}
\begin{document}
\maketitle
\companyauthornotes
\section{Introduction}
\label{sec:introduction}

Autoregressive vision-language-action (VLA) models generate discrete tokens that are decoded into executable robot commands~\citep{RT2_23_arxiv,OpenVLA_24,FAST_25}. The action tokenizer connects continuous demonstrations to this discrete prediction process: it determines both the targets used for policy training and the commands recovered from predicted tokens. Existing approaches construct this interface through coordinate discretization, frequency-domain compression, or learned vector quantization~\citep{OpenVLA_24,FAST_25,dong2026actioncodec,liu2025faster,liu2026oat,liu2026oatpolicy}. These designs enable compact action generation, but compression also changes the actions that the robot ultimately executes. An important question is therefore what an action tokenizer should preserve to support precise control.

Across multiple demonstrations of the same task, similar motions recur with adjustments to object positions, robot states, and execution stages.
These adjustments allow similar motions to accommodate different contexts. These adjustments allow similar motions to accommodate different contexts. In operations requiring precise alignment, grasping, or contact, even small action differences can affect execution outcomes. A tokenizer compresses these demonstrations into a discrete representation with finite capacity and reconstructs executable commands from it. Although reconstruction objectives such as mean squared error (MSE) keep individual actions close to their originals, they do not explicitly coordinate reconstruction errors across demonstrations. Small individual reconstruction errors therefore do not fully characterize how faithfully the adjustments between demonstrations are preserved. After compression, similar actions may still cluster around a representative motion, while the adjustments needed to accommodate different contexts are distorted, diminished, or even reversed. These changes persist even when a policy correctly predicts the corresponding token sequence: the decoder still executes the reconstructed action.

This motivates studying \textbf{relational fidelity alongside pointwise reconstruction accuracy}. We characterize action differences using a physical distance that combines translation, rotation, and gripper state, and examine whether their relative magnitudes survive tokenization. To measure this property, we introduce \textbf{physical rank consistency} (PRC). For each action chunk, PRC compares its distance rankings to the same original neighbors before and after reconstruction. Computing these rankings in decoded action space provides a common physical reference across token vocabularies, sequence lengths, and decoder architectures. Together, reconstruction accuracy and PRC describe how faithfully a tokenizer reproduces individual commands and preserves the local relationships among them.

To preserve these relationships during tokenization, we develop \methodname. Physical distances between original action chunks provide near--far supervision for both representation learning and codeword assignment. \textbf{Physical rank preservation} (PRP) encourages the corresponding ordering in a combination of encoder and quantized feature distances. \textbf{Quantization regularization} (QR) applies this ordering to the divergence between codeword assignment distributions, extending supervision to how actions are assigned to the discrete vocabulary. Together, these objectives address two stages of compression: learning action features and mapping them to codewords. They are optimized jointly with reconstruction in a Transformer tokenizer with residual vector quantization. After tokenizer training, the frozen encoder and quantizer supply discrete policy targets, while the frozen decoder converts autoregressively predicted tokens into executable action chunks.

Under a shared Qwen3-VL-4B policy training setup, \methodname\ outperforms all compared action tokenizers, achieving \liberoOurs\% success on LIBERO and \plusOurs\% on unseen LIBERO-Plus, exceeding the strongest baseline on each benchmark by 1.1 and 4.5 percentage points, respectively (Table~\ref{tab:libero-tokenizer-comparison}). Evaluations on SimplerEnv and VLA-Arena extend the comparison to real-to-sim transfer and challenging control conditions, where \methodname\ achieves state-of-the-art aggregate success rates of \simplerOurs\% and \vlaMean\%, respectively, among the evaluated methods. These results are obtained through standard next-token prediction using \methodname\ as the action tokenizer. Controlled ablations show that PRP and QR jointly improve both decoded physical rank consistency and downstream policy success. Comparisons across tokenizers further examine the relationship between reconstruction fidelity, PRC, and execution performance (Figure~\ref{fig:proxy-validation}). 

Our contributions are threefold:
\begin{itemize}
    \item We introduce PRC, a measure of local physical distance-order
    preservation that complements reconstruction accuracy and supports
    comparison across action tokenizers.
    \item We develop \methodname, jointly supervising physical relationships
    in learned features and codeword assignment distributions to construct
    discrete action representations for autoregressive VLA policies.
    \item We evaluate \methodname\ across four benchmarks and conduct
    controlled comparisons and ablations to examine the effects of physical
    relationship supervision on representation quality and robot execution.
\end{itemize}

\section{Related Work}
\label{sec:related}

\subsection{Action-Space Design in Robotic Foundation Models}

Robot foundation models connect pretrained vision--language representations to motor commands through discrete, continuous, or hybrid action interfaces. Discrete autoregressive approaches cast control as next-token prediction: RT-2 and OpenVLA discretize action coordinates into bins ~\citep{RT2_23_arxiv,OpenVLA_24}, while FAST compresses action chunks into shorter token sequences~\citep{FAST_25}. VLM2VLA instead expresses commands using the VLM's existing natural-language vocabulary ~\citep{VLM2VLA_25}. These interfaces retain the token-prediction objective of pretrained VLMs, providing a direct route to transfer language grounding and semantic knowledge to robot control.

Continuous approaches predict real-valued action chunks without categorical decoding. OpenVLA-OFT uses parallel action prediction with an $\ell_1$ regression objective~\citep{OpenVLA-OFT_2025}; diffusion and flow-matching policies, including RDT, $\pi_0$, and the GR00T family, model continuous action distributions~\citep{RDT_25_ICLR,PI0,GR00T_25,GR00T_N1.6}. Continuous heads support fine-grained control, and generative heads can represent multiple valid motions while generating action chunks without autoregressively decoding every action token.

Hybrid designs combine discrete supervision with continuous execution. $\pi_{0.5}$ uses action tokens during pretraining and introduces a flow-matching action expert during post-training~\citep{PI05_25}. Knowledge Insulation trains the backbone with discrete action targets while blocking gradients from the continuous expert, retaining pretrained knowledge alongside efficient execution~\citep{knowledgeinsulation2025}. HybridVLA jointly trains autoregressive and diffusion predictions and adaptively combines them~\citep{HyBridVLA_25}; Fast-in-Slow couples autoregressive supervision with a fast diffusion-based execution module ~\citep{fastinslow2025}. Such designs seek to combine the semantic benefits of token prediction with responsive continuous control.

\subsection{Action Discretization and Tokenization}

Coordinate-wise binning provides a simple action vocabulary, as in RT-2 and OpenVLA~\citep{RT2_23_arxiv,OpenVLA_24}, but represents each coordinate and time step separately. FAST applies a discrete cosine transform to action chunks, quantizes the coefficients, and learns byte-pair encoding (BPE) to compress the resulting sequences~\citep{FAST_25}. Its frequency basis is fixed, while the BPE vocabulary is learned from action data.

Learned latent tokenizers instead optimize an encoder and decoder around a quantized representation. Vector-quantized approaches learn codebooks from action data; FASTer combines residual vector quantization (RVQ), structured action patching, temporal and spectral reconstruction, and blockwise autoregressive generation~\citep{liu2025faster}. OAT uses finite scalar quantization with nested dropout and causal attention to obtain compact, fully decodable tokens with coarse-to-fine ordering ~\citep{liu2026oat,liu2026oatpolicy}. ActionCodec studies overlap, token budget, vision--language alignment, and residual grammar from the perspective of VLA optimization~\citep{dong2026actioncodec}. X-Tokenizer adds semantic supervision from a frozen visual--language teacher~\citep{kang2026xtokenizer}. These methods develop action interfaces through compression, token ordering, and multimodal alignment. ActionPiece introduces explicit physical-order supervision during encoding and quantization, and uses PRC to measure neighborhood-order preservation in decoded action space.
\section{Physical Structure in Action Tokenization}
\label{sec:motivation}

\subsection{Preserving Action Variations}

Demonstrations of the same task contain recurring motions with adjustments to object positions, robot states, and execution stages. These adjustments allow similar motions to accommodate different contexts. In operations requiring precise alignment, grasping, or contact, even small action differences can affect execution outcomes.

An action tokenizer compresses continuous action chunks into discrete token sequences with finite representational capacity, then decodes them into executable commands. This lossy compression introduces reconstruction errors. Although tokenizers are commonly trained with an MSE objective to keep reconstructed actions close to their originals, this objective penalizes individual reconstruction errors without explicitly coordinating their directions across actions. Consider the normalized Euclidean components $x_i,x_j$ of two demonstration chunks, with reconstruction errors $e_i=\widehat x_i-x_i$. Their reconstructed difference satisfies \begin{equation} \widehat x_j-\widehat x_i =(x_j-x_i)+(e_j-e_i). \label{eq:relative-reconstruction} \end{equation} The first term is the action variation present in the demonstrations; the second is the change introduced by tokenization. This additional term can attenuate the original difference, amplify it, or even reverse its direction. Small individual reconstruction errors therefore do not fully characterize how faithfully the adjustments between demonstrations are preserved. After compression, similar actions may still cluster around a representative motion, while their original near-to-far ordering is altered.

These distortions carry through to policy execution. Even when a VLA correctly predicts the token sequence encoding a demonstrated action, the decoder still produces its reconstructed version. If compression weakens or distorts the adjustments required by different contexts, correct token prediction cannot recover the original adjustments. Action tokenization should therefore preserve physical distinctions among demonstrations alongside accurate reconstruction of individual actions.

\subsection{Physical Rank Consistency}

We assess relationship preservation through the relative ordering of physical distances between action chunks. Our distance $d_{\mathrm{phys}}$ combines translation, shortest geodesic rotation on $\mathrm{SO}(3)$, and gripper differences across a chunk, as defined in Section~\ref{sec:physical-distance}. For each anchor $A_i$, let $\mathcal N_i^k$ contain its $k$ nearest neighbors in the original action space, and let $\widehat A_i=\mathcal D(\mathcal T(A_i))$. Physical Rank Consistency (PRC) compares distances to these same neighbors before and after reconstruction: \begin{equation} \operatorname{PRC@}k =\frac{1}{N}\sum_{i=1}^{N}\rho_{\mathrm S}\!\left( [d_{\mathrm{phys}}(A_i,A_j)]_{j\in\mathcal N_i^k}, [d_{\mathrm{phys}}(\widehat A_i,\widehat A_j)]_{j\in\mathcal N_i^k} \right), \label{eq:PRC} \end{equation} where $\rho_{\mathrm S}$ denotes Spearman correlation. We use $k=32$.

Higher PRC indicates better preservation of distance ordering within the original neighborhoods. Comparing ranks captures the relative size of action differences without requiring every distance to remain numerically identical. Evaluation in decoded action space provides a common physical reference across token vocabularies, sequence lengths, and decoder architectures. MSE measures the fidelity of individual reconstructions, while PRC measures the preservation of physical distance ordering. Figure~\ref{fig:proxy-validation} examines both properties in relation to downstream policy success. Guided by this perspective, ActionPiece introduces physical-order supervision into representation learning and codeword assignments, preserving action relationships alongside pointwise reconstruction.
\section{ActionPiece}
\label{sec:method}

\begin{figure}[t]
  \centering
  \includegraphics[width=\textwidth]{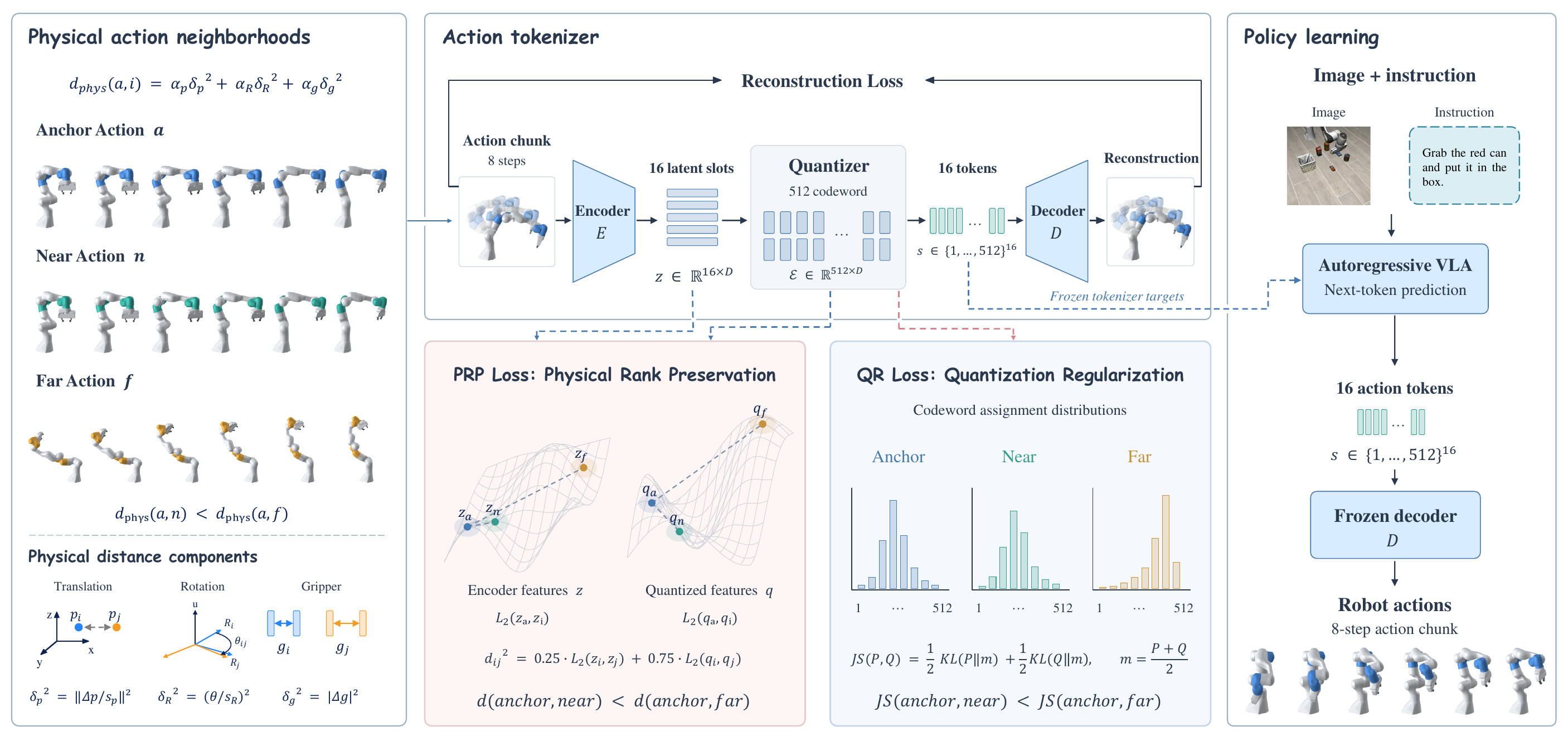}
  \caption{\textbf{Overview of ActionPiece.}
  Physical action distance (defined in Section~\ref{sec:physical-distance}) combines normalized translation, shortest-path rotation on $\mathrm{SO}(3)$, and gripper differences to select anchor, near, and far action chunks within a training batch. The tokenizer compresses each eight-step chunk into 16 discrete tokens and reconstructs executable actions. Alongside reconstruction, physical rank preservation (PRP, defined in Section~\ref{sec:rank-preservation}) encourages the near--far ordering in encoder and quantized feature distances, while quantization regularization (QR, defined in Section~\ref{sec:quantization-regularization}) encourages the same ordering in the Jensen--Shannon divergence between codeword assignment distributions. Here $L_2$ denotes the slot-averaged squared distance between normalized features. }
  \label{fig:architecture}
\end{figure}

ActionPiece preserves physical action relationships alongside reconstruction accuracy through joint supervision of encoding and quantization, following Section~\ref{sec:motivation}. A shallow Transformer encoder and decoder with residual vector quantization (RVQ) provide the codec. Physical rank preservation structures the learned action representations; quantization regularization carries the same ordering into codeword assignments. Both augment reconstruction during tokenizer training. The resulting vocabulary supplies the targets for standard autoregressive policy learning. Figure~\ref{fig:architecture} summarizes the tokenizer training objectives and the resulting policy learning and execution pipeline.

\subsection{Compact, Fully Decodable Tokens}

For an action chunk $A=(a_1,\ldots,a_H)\in\mathbb{R}^{H\times D}$, the encoder $E_\phi$ produces $S$ latent slots. An RVQ with depth $Q$ and $K$ entries per codebook represents slot $s$ as \begin{equation} q_s=\sum_{r=1}^{Q}e^{(r)}[z_s^{(r)}], \qquad Z=\{z_s^{(r)}\}_{s=1,r=1}^{S,Q}, \qquad |Z|=SQ=T. \label{eq:rvq-tokenization} \end{equation} The decoder reconstructs the complete chunk in one pass, $\widehat A=D_\omega(q)$. Here $T$ is the number of categorical outputs generated by the VLA, while $Q$ controls how the budget is divided between latent slots and residual levels. Every code sequence maps to a complete action chunk.

\subsection{Physical Action Distance}
\label{sec:physical-distance}

For two end-effector commands $a=(p,R,g)$ and $a'=(p',R',g')$, we measure rotational distance by the shortest geodesic angle on $\mathrm{SO}(3)$: \begin{equation} d_{\mathrm{SO}(3)}(R,R') =\left\|\operatorname{Log}(R^\top R')^{\vee}\right\|_2, \end{equation} where $\operatorname{Log}(\cdot)^{\vee}$ is the rotation-vector form of the principal matrix logarithm. The per-step physical distance combines translation, rotation, and gripper state: \begin{equation} \delta_{\mathrm{phys}}(a,a') =\alpha_p\left\|\frac{p-p'}{s_p}\right\|_2^2 +\alpha_R\left(\frac{d_{\mathrm{SO}(3)}(R,R')}{s_R}\right)^2 +\alpha_g|g-g'|^2. \label{eq:physical-distance} \end{equation} The scales are fitted on training actions, and the group weights are fixed within an embodiment. For action chunks, $d_{\mathrm{phys}}$ averages these per-step distances over corresponding time steps. Rotations are mapped to valid elements of $\mathrm{SO}(3)$ before computing the distance. This distance defines the near-to-far ordering used by the two training objectives and by PRC in decoded action space.

\subsection{Physical Rank Preservation}
\label{sec:rank-preservation}

Physical rank preservation aligns the relative distances of learned action representations with those of physical motions. We measure representation distance both before and after quantization. Let $\bar h_{i,s}$ be the encoder representation of action chunk $A_i$ at slot $s$, and let $\bar e_{i,s}$ be its selected codeword after the quantizer output projection. Both are normalized by LayerNorm followed by unit $\ell_2$ normalization. Their distances are \begin{equation} \begin{aligned} d_{\mathrm{enc}}^2(i,j) &=\frac{1}{S}\sum_{s=1}^{S} \|\bar h_{i,s}-\bar h_{j,s}\|_2^2,\\ d_{\mathrm{quant}}^2(i,j) &=\frac{1}{S}\sum_{s=1}^{S} \|\bar e_{i,s}-\bar e_{j,s}\|_2^2. \end{aligned} \label{eq:quantized-geometry} \end{equation} We use the weighted representation distance \begin{equation} d_{\mathrm{rep}}^2(i,j) =0.25\,d_{\mathrm{enc}}^2(i,j) +0.75\,d_{\mathrm{quant}}^2(i,j), \qquad D_{ij}=\sqrt{d_{\mathrm{rep}}^2(i,j)}. \label{eq:representation-distance} \end{equation} This distance incorporates the encoder features and the discrete representations used by the decoder. A differentiable codeword estimator transmits gradients through code selection; Appendix~\ref{app:quantization-estimator} specifies the estimator, and Appendix~\ref{app:multiple-levels} covers multiple quantization levels.

For each anchor action chunk $A_i$, we exclude the anchor itself and rank the remaining action chunks in the training batch by physical distance. We select the positive (near) action $A_{j_i^+}$ at the 5th percentile ($q_{0.05}$) and the negative (far) action $A_{j_i^-}$ at the 95th percentile ($q_{0.95}$). The rank objective is \begin{equation} \mathcal L_{\mathrm{rank}} =\frac{1}{B}\sum_i \operatorname{softplus}\!\left(D_{i j_i^+}-D_{i j_i^-}+m_r\right), \qquad m_r=0.1. \label{eq:rank} \end{equation} The loss encourages representations to retain the near-to-far order of physical motions. Both physical-order objectives use these pairs; Appendix~\ref{app:neighbor-sampling} specifies the selection rule.

\subsection{Quantization Regularization}
\label{sec:quantization-regularization}

Quantization regularization applies physical-order supervision directly to codeword probabilities. Let $\pi_{i,s}$ denote the soft probability distribution over codewords for slot $s$ of action chunk $A_i$, obtained from encoder-to-codeword distances as specified in Eq.~\eqref{eq:codeword-probabilities}. We compare two action chunks using the average Jensen--Shannon divergence between corresponding slots: \begin{equation} d_{\mathrm{JS}}(i,j) =\frac{1}{S}\sum_{s=1}^{S} \operatorname{JS}(\pi_{i,s},\pi_{j,s}), \label{eq:probability-distance} \end{equation} where $\operatorname{JS}(p,q)=\tfrac12\operatorname{KL}(p\|\mu) +\tfrac12\operatorname{KL}(q\|\mu)$ and $\mu=(p+q)/2$. Using the same neighbors $j_i^+$ and $j_i^-$ as the rank objective, we define \begin{equation} \mathcal L_{\mathrm{quant}} =\frac{1}{B}\sum_i \operatorname{softplus}\!\left( d_{\mathrm{JS}}(i,j_i^+)-d_{\mathrm{JS}}(i,j_i^-)+m_q\right), \qquad m_q=0.1. \label{eq:quantization-regularization} \end{equation} Physical rank preservation supervises distances between action representations, while quantization regularization supervises the probabilities that determine codeword selection. Together, they impose physical order on these two aspects of the tokenizer.

\subsection{Training and Policy Integration}

The reconstruction objective measures mean squared error in normalized action coordinates: \begin{equation} \mathcal L_{\mathrm{MSE}} =\frac{1}{BHD}\sum_{i,t,d} \left(\frac{A_{i,t,d}-\widehat A_{i,t,d}}{s_d}\right)^2. \label{eq:mse-reconstruction} \end{equation} The complete tokenizer objective combines reconstruction, the standard codebook commitment loss, and the two physical-order objectives: \begin{equation} \mathcal L_{\mathrm{tok}} =\mathcal L_{\mathrm{MSE}} +\lambda_{\mathrm{vq}}\mathcal L_{\mathrm{commit}} +\lambda_r\mathcal L_{\mathrm{rank}} +\lambda_q\mathcal L_{\mathrm{quant}}. \label{eq:tokenizer-objective} \end{equation} We set $\lambda_{\mathrm{vq}}=0.25$ and $\lambda_r=\lambda_q=6.25\times10^{-4}$. The component ablation in Table~\ref{tab:component-ablation} removes either physical-order objective or both from this formulation. PRC evaluates the physical neighborhood structure of decoded actions after training.

After training, the tokenizer is frozen and its indices are added to the VLM vocabulary. The VLA is optimized with standard next-token prediction, \begin{equation} \mathcal L_{\mathrm{VLA}} =-\sum_{t=1}^{T}\log p_\theta(z_t\mid I,\ell,z_{<t}), \label{eq:vla-objective} \end{equation} and the generated sequence is decoded into an executable action chunk.
\section{Experiments}
\label{sec:experiments}

We relate tokenizer fidelity and physical structure to policy success, then
test ActionPiece through controlled comparisons, transfer evaluations, and ablations.

\begin{figure*}[t]
    \centering
    \includegraphics[width=\textwidth]{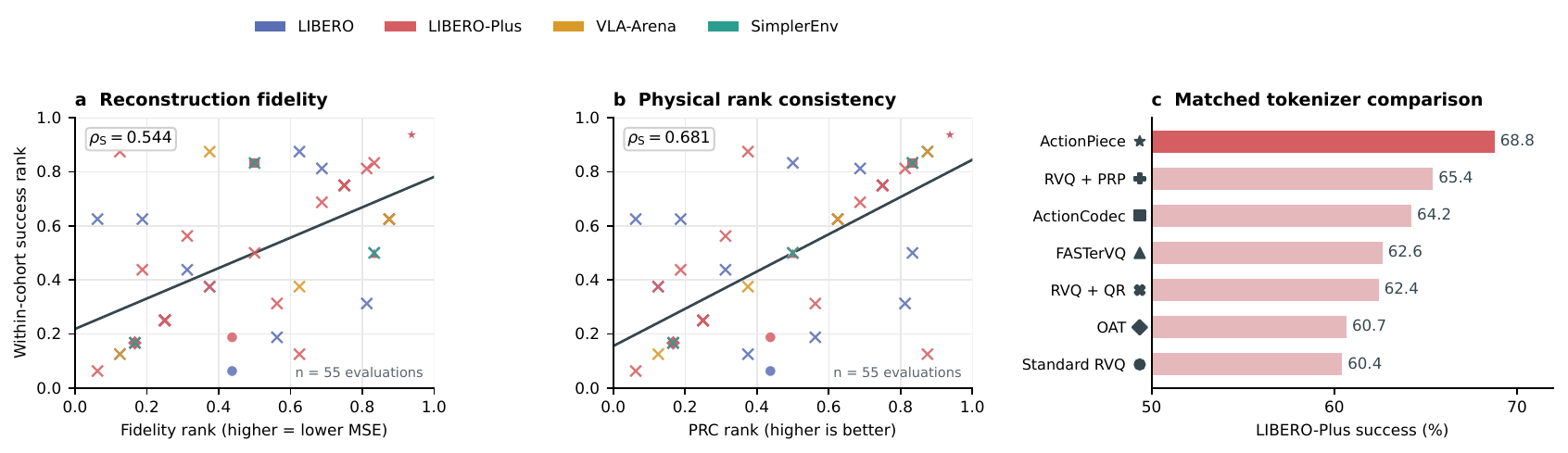}
    \caption{\textbf{Reconstruction fidelity, physical structure, and policy success.}
    Panels (a--b) show normalized within-group ranks across 55
    tokenizer--benchmark evaluations. Groups follow the original experimental
    batches, with benchmarks evaluated separately. Colors indicate benchmarks;
    symbols identify the methods shown in (c), and crosses denote additional
    variants. Panel (c) reports LIBERO-Plus success for selected tokenizers and
    ablations using Qwen3-VL-4B at 30K policy training steps. PRP and QR denote
    physical rank preservation and quantization regularization.}
    \label{fig:proxy-validation}
\end{figure*}

\subsection{Experimental Protocol}

\paragraph{Benchmarks and data.}
We select four benchmarks spanning three action-data sources and complementary
generalization settings. LIBERO~\citep{libero} provides the in-distribution
reference, using human teleoperation demonstrations collected in simulation.
LIBERO-Plus~\citep{fei25libero-plus} evaluates these LIBERO-trained policies
under seven types of unseen perturbations, without training on LIBERO-Plus data.
SimplerEnv~\citep{SimplerEnv_24} evaluates a WidowX policy trained on real-robot
BridgeData V2 demonstrations~\citep{walke2023bridgedata}, testing real-to-sim
transfer. For VLA-Arena~\citep{VLA-Arena_25}, we use keyboard-teleoperated
simulation demonstrations at L0 and evaluate across L0--L2, with L1/L2 testing
generalization to more demanding task configurations. Together, these settings
test tokenizers across different demonstration sources and assess policy
performance both within and beyond the training distribution.

\paragraph{Tokenizer training.}
Action data are sampled at 20\,Hz for LIBERO, 10\,Hz for VLA-Arena, and 5\,Hz for BridgeData V2. Tokenizers are trained on action chunks using AdamW~\citep{adamw_2017} with a learning rate of $1\times10^{-4}$ and a batch size of 128 for 100K steps. For fair comparison, each baseline tokenizer is trained using the remaining optimizer hyperparameters specified in its original work. Within each controlled comparison, the action representation, normalization, and execution settings are fixed.

\paragraph{VLA training.}
All experiments use eight NVIDIA RTX PRO 6000 GPUs. For policy training, we follow the default StarVLA protocol~\citep{starVLA_26_arxiv} and use AdamW~\citep{adamw_2017} with an initial learning rate of $1\times10^{-5}$ and a cosine annealing schedule. We use DeepSpeed ZeRO-2~\citep{deepspeed_2020}, gradient clipping with a maximum norm of 1.0, and no gradient accumulation.

\paragraph{Matched policy evaluation.}
Within each controlled comparison, the VLM backbone, prompt, robot demonstrations, optimizer, global batch, training budget, action and execution horizons, seed, and evaluation protocol are fixed. Full LIBERO contains 2,000 rollouts, full LIBERO-Plus 10,030, and complete VLA-Arena 3,400.

% Matched action-tokenizer comparison on full LIBERO and LIBERO-Plus.
\begin{table*}[!t]
\centering
\begingroup
\papertablestyle
\setlength{\tabcolsep}{3.2pt}
\caption{\textbf{Matched action-tokenizer comparison on LIBERO and LIBERO-Plus.}
All policies use the same Qwen3-VL-4B backbone, training data, prompt,
vision--language co-training procedure, global batch size 128,
an 8-step prediction and execution horizon, and maximum policy-training budget.
LIBERO-Plus is evaluated out of distribution: no LIBERO-Plus demonstrations
are used for policy training. FASTerVQ$^*$ denotes our implementation following
the published method. Best results are bold.}
\label{tab:libero-tokenizer-comparison}
\resizebox{\textwidth}{!}{%
\begin{tabular}{@{}lccccc@{\hspace{8pt}}cccccccc@{}}
\toprule
& \multicolumn{5}{c}{\textbf{LIBERO}}
& \multicolumn{8}{c}{\textbf{LIBERO-Plus}} \\
\cmidrule(lr){2-6}\cmidrule(lr){7-14}
\textbf{Tokenizer} & Spatial & Object & Goal & Long & Overall
& Camera & Robot & Language & Light & Bkg. & Noise & Layout & Overall \\
\midrule
ActionCodec~\citep{dong2026actioncodec}
& 93.4 & \textbf{99.0} & 92.4 & 89.8 & 93.7
& 33.3 & 38.8 & 80.8 & 88.4 & 90.2 & \textbf{58.0} & 75.9 & 64.2 \\
OAT~\citep{liu2026oat,liu2026oatpolicy}
& 85.6 & 96.2 & 87.2 & 75.2 & 86.1
& 32.2 & 47.8 & 79.5 & 82.8 & 82.7 & 46.4 & 67.4 & 60.7 \\
FASTerVQ$^*$~\citep{liu2025faster}
& 91.8 & 95.6 & 90.5 & 87.4 & 91.3
& 39.0 & 42.3 & 74.5 & 85.8 & 84.9 & 56.6 & 69.0 & 62.6 \\
FAST~\citep{FAST_25}
& 94.7 & 98.0 & 93.1 & 82.6 & 92.1
& 33.8 & 47.1 & 80.0 & 89.6 & 88.4 & 51.3 & 75.9 & 64.3 \\
Standard RVQ
& 91.8 & 98.2 & 91.6 & 82.0 & 90.9
& 31.5 & 43.9 & 75.3 & 86.0 & 88.0 & 42.5 & 72.9 & 60.4 \\
\rowcolor{gray!30}\textbf{\methodname (Ours)}
& \textbf{96.2} & \textbf{99.0} & \textbf{93.4} & \textbf{90.6} & \textbf{94.8}
& \textbf{45.1} & \textbf{49.6} & \textbf{82.2} & \textbf{91.9}
& \textbf{91.4} & 57.6 & \textbf{78.0} & \textbf{68.8} \\
\bottomrule
\end{tabular}%
}
\endgroup
\end{table*}

\subsection{Reconstruction and Physical Structure}
\label{sec:proxy-validation}

Figure~\ref{fig:proxy-validation} examines how tokenizer reconstruction fidelity and neighborhood-order preservation relate to downstream policy success, examining the action-space property introduced in Section~\ref{sec:motivation}. We compute normalized ranks within the original experimental groups and summarize their association across 55 tokenizer--benchmark evaluations. Both metrics exhibit positive associations with success, with PRC yielding a higher Spearman correlation ($0.681$) than reconstruction fidelity ($0.544$). This observation motivates examining physical structure alongside pointwise reconstruction; the controlled comparisons and ablations below evaluate the benefits of explicitly supervising this structure. Appendix~\ref{app:tokenizer-variants} describes alternative RVQ objectives and reports their reconstruction, PRC, and policy success.

\subsection{Matched Action-Tokenizer Comparison}

We compare ActionPiece with ActionCodec~\citep{dong2026actioncodec}, OAT
~\citep{liu2026oat,liu2026oatpolicy}, FASTer~\citep{liu2025faster}, FAST~\citep{FAST_25}, and standard RVQ through one
Qwen3-VL-4B policy implementation. The policy data, prompt, optimizer, budget, and
evaluation are shared. Each tokenizer retains its native output length and vocabulary. The comparison
evaluates the policy trained with each action representation.

Table~\ref{tab:libero-tokenizer-comparison} reports \liberoOurs\% on in-distribution LIBERO for \methodname, compared with \liberoBestBaseline\% for ActionCodec, the strongest baseline on this benchmark. The separation is larger on unseen LIBERO-Plus: \methodname reaches \plusOurs\%, compared with \plusBestBaseline\% for the strongest baseline, and leads on six of seven shift categories. All policies are trained on LIBERO demonstrations; LIBERO-Plus measures transfer under camera, robot, language, illumination, background, noise, and layout changes. The advantage over the strongest baseline grows from 1.1 points on LIBERO to 4.5 on LIBERO-Plus. Gains across six shift categories support preserving action adjustments as a useful principle for execution beyond the training distribution.

\FloatBarrier
\subsection{Evaluation on SimplerEnv and VLA-Arena}

In both benchmarks, we integrate ActionPiece by replacing the action tokenizer
while retaining the policy architecture and training protocol. Policies are
trained with standard next-token prediction, without benchmark-specific
policy modifications.

% SimplerEnv WidowX comparison under the benchmark's repeated-trial protocol.
\begin{table}[!t]
\centering
\begingroup
\papertablestyle
\caption{\textbf{SimplerEnv WidowX results.}
We follow the official SimplerEnv evaluation protocol~\citep{SimplerEnv_24}. All values are success rates
(\%). Best and second-best averages are bold and underlined, respectively.}
\label{tab:simplerenv}
\begin{adjustbox}{max width=\linewidth}
\begin{tabular}{@{}lccccc@{}}
\toprule
\textbf{Method}
& \shortstack{\textbf{Put Spoon}\\on Towel}
& \shortstack{\textbf{Put Carrot}\\on Plate}
& \shortstack{\textbf{Stack Block}\\on Block}
& \shortstack{\textbf{Put Eggplant}\\in Basket}
& \textbf{Avg.} \\
\midrule
RT-1-X~\citep{OXE_24} & 0.0 & 4.2 & 0.0 & 0.0 & 1.1 \\
Octo-Base~\citep{Octo_2024} & 0.0 & 12.5 & 15.8 & 41.7 & 17.5 \\
Octo-Small~\citep{Octo_2024} & 0.0 & 8.2 & 41.7 & 56.7 & 26.7 \\
OpenVLA-OFT~\citep{OpenVLA-OFT_2025} & 34.2 & 30.0 & 30.0 & 72.5 & 41.8 \\
RoboVLM~\citep{RoboVLM_2024} & 50.0 & 37.5 & 0.0 & 83.3 & 42.7 \\
Magma~\citep{yang2025magma} & 37.5 & 29.2 & 20.8 & 91.7 & 44.8 \\
CogACT~\citep{CogACT_2024} & 71.7 & 50.8 & 15.0 & 67.5 & 51.3 \\
SpatialVLA~\citep{SpatialVLA_2025} & 20.8 & 20.8 & 25.0 & 70.8 & 34.4 \\
TraceVLA~\citep{TraceVLA_2025} & 12.5 & 16.6 & 16.6 & 65.0 & 27.7 \\
VideoVLA~\citep{VideoVLA_2025} & 75.0 & 20.8 & 45.8 & 70.8 & 53.1 \\
$\pi_0$~\citep{PI0} & 29.2 & 62.5 & 29.2 & 91.6 & 53.1 \\
$\pi_{0.5}$~\citep{PI05_25} & 49.3 & 64.7 & 44.7 & 69.7 & 57.1 \\
Isaac-GR00T-N1.6-Bridge~\citep{GR00T_N1.6} & 64.5 & 65.5 & 5.5 & 93.0 & 57.1 \\
LangForce~\citep{LangForce} & 89.6 & 63.8 & 33.3 & 79.2 & \underline{66.5} \\
ActionCodec-BAR~\citep{dong2026actioncodec} & 71.7 & 64.2 & 55.0 & 70.0 & 65.2 \\
\rowcolor{gray!30}\textbf{\methodname (Ours)} & 83.3 & 54.2 & 58.3 & 91.7 &
\textbf{\simplerOurs} \\
\bottomrule
\end{tabular}
\end{adjustbox}
\endgroup
\end{table}

For each SimplerEnv task, we average success over five independent
24-episode repeats, then macro-average the four task scores.
ActionPiece achieves the highest average success of \simplerOurs\% in this
real-to-sim evaluation, with task-specific strengths varying across methods.
ActionPiece leads block stacking at 58.3\%, a task requiring precise placement,
alongside strong spoon (83.3\%) and eggplant (91.7\%) results.
These outcomes extend the evidence for physical relationship supervision to
tokenizers trained on real-robot actions.

% Success rates in percent; cumulative costs retain their original units.
\begin{table*}[!t]
\centering
\begingroup
\papertablestyle
\setlength{\tabcolsep}{2.5pt}
\renewcommand{\arraystretch}{1.1}
\caption{\textbf{VLA-Arena results}~\citep{VLA-Arena_25}. SR is success rate (\%); CC is cumulative
cost (lower is better). Avg equally weights the 11 suites. For each task and difficulty level, the highest SR across the six models is bold,
including ties; the same rule applies to Avg. CC values are not bold. Baselines include $\pi_{0.5}$~\citep{PI05_25}, GR00T-N1.6~\citep{GR00T_N1.6}, Evo-Depth~\citep{lin2026evodepth}, and Motus~\citep{bi2025motusunifiedlatentaction}. Qwen-GR00T denotes Qwen3-VL~\citep{Qwen3-VL} with the GR00T N1 action head~\citep{GR00T_25}.}
\label{tab:vla-arena}
\begin{adjustbox}{max width=\linewidth}
\begin{tabular}{@{}ll*{6}{ccc}@{}}
\toprule
\multirow{2}{*}{\textbf{Task}} & \multirow{2}{*}{\textbf{Metric}} & \multicolumn{3}{c}{$\pi_{0.5}$} & \multicolumn{3}{c}{GR00T-N1.6} & \multicolumn{3}{c}{Qwen-GR00T} & \multicolumn{3}{c}{Evo-Depth} & \multicolumn{3}{c}{Motus} & \multicolumn{3}{c}{\textbf{\methodname}} \\
\cmidrule(lr){3-5}\cmidrule(lr){6-8}\cmidrule(lr){9-11}\cmidrule(lr){12-14}\cmidrule(lr){15-17}\cmidrule(lr){18-20}
& & L0 & L1 & L2 & L0 & L1 & L2 & L0 & L1 & L2 & L0 & L1 & L2 & L0 & L1 & L2 & L0 & L1 & L2 \\
\midrule
\rowcolor{gray!15}\multicolumn{20}{l}{\textbf{Safety}} \\
\multirow{2}{*}{Static obstacles} & SR $\uparrow$ & 90.0 & 62.0 & 40.0 & 72.0 & 30.0 & 14.0 & \textbf{91.0} & 18.0 & 5.0 & 88.0 & 66.0 & 48.0 & 64.0 & 67.0 & 41.0 & \textbf{91.0} & \textbf{83.0} & \textbf{57.0} \\
 & CC $\downarrow$ & 0.0 & 33.3 & 76.6 & 0.0 & 11.2 & 38.4 & 0.0 & 9.1 & 29.6 & 0.0 & 8.7 & 14.9 & 0.0 & 72.0 & 124.3 & 0.0 & 14.5 & 30.3 \\
\addlinespace
\multirow{2}{*}{Cautious grasp} & SR $\uparrow$ & 50.0 & 14.0 & 0.0 & 16.0 & 2.0 & 0.0 & 77.0 & 8.0 & 2.0 & \textbf{78.0} & 24.0 & 0.0 & 62.0 & \textbf{25.0} & \textbf{11.0} & 77.0 & 18.0 & 1.0 \\
 & CC $\downarrow$ & 5.0 & 5.5 & 1.2 & 9.6 & 41.2 & 10.4 & 2.4 & 98.2 & 11.1 & 0.0 & 11.4 & 26.3 & 3.9 & 91.0 & 9.8 & 1.1 & 19.3 & 12.4 \\
\addlinespace
\multirow{2}{*}{Hazard avoidance} & SR $\uparrow$ & 58.0 & 30.0 & 36.0 & 64.0 & 4.0 & 10.0 & 67.0 & 15.0 & 19.0 & 40.0 & 0.0 & 14.0 & 65.0 & \textbf{34.0} & 21.0 & \textbf{68.0} & 22.0 & \textbf{46.0} \\
 & CC $\downarrow$ & 7.1 & 15.0 & 14.5 & 6.1 & 20.2 & 17.9 & 7.0 & 19.3 & 19.5 & 7.4 & 25.9 & 7.9 & 15.6 & 40.2 & 51.3 & 0.7 & 4.0 & 1.7 \\
\addlinespace
\multirow{2}{*}{State preservation} & SR $\uparrow$ & 58.0 & 56.0 & 54.0 & 66.0 & 50.0 & 38.0 & 86.0 & 56.0 & 47.0 & 88.0 & 66.0 & 56.0 & 63.0 & 63.0 & \textbf{72.0} & \textbf{94.0} & \textbf{80.0} & 64.0 \\
 & CC $\downarrow$ & 0.0 & 5.6 & 20.8 & 0.0 & 5.0 & 10.4 & 0.0 & 5.6 & 15.7 & 0.0 & 2.3 & 5.1 & 0.0 & 6.3 & 16.0 & 0.0 & 6.9 & 25.3 \\
\addlinespace
\multirow{2}{*}{Dynamic obstacles} & SR $\uparrow$ & 50.0 & 44.0 & 22.0 & 74.0 & 50.0 & 2.0 & 81.0 & 56.0 & 3.0 & \textbf{82.0} & 60.0 & 6.0 & 58.0 & 33.0 & 11.0 & \textbf{82.0} & \textbf{76.0} & \textbf{27.0} \\
 & CC $\downarrow$ & 2.4 & 8.8 & 5.7 & 5.7 & 7.3 & 56.8 & 6.0 & 8.3 & 2.7 & 0.0 & 3.4 & 12.7 & 6.5 & 47.9 & 7.2 & 4.6 & 21.2 & 31.0 \\
\addlinespace
\midrule
\rowcolor{gray!15}\multicolumn{20}{l}{\textbf{Distractor}} \\
Static distractors & SR $\uparrow$ & 88.0 & 16.0 & 16.0 & 46.0 & \textbf{32.0} & 6.0 & 91.0 & 6.0 & 2.0 & \textbf{94.0} & 20.0 & \textbf{24.0} & 69.0 & 19.0 & 13.0 & 92.0 & 21.0 & 22.0 \\
Dynamic distractors & SR $\uparrow$ & 80.0 & 66.0 & \textbf{54.0} & 70.0 & \textbf{72.0} & 18.0 & \textbf{92.0} & 49.0 & 25.0 & 86.0 & 60.0 & 32.0 & 77.0 & 51.0 & 26.0 & 90.0 & 63.0 & 43.0 \\
\midrule
\rowcolor{gray!15}\multicolumn{20}{l}{\textbf{Extrapolation}} \\
\mbox{Prep. combinations} & SR $\uparrow$ & 62.0 & \textbf{24.0} & \textbf{6.0} & 48.0 & 0.0 & 0.0 & 51.0 & 1.0 & 0.0 & 66.0 & 0.0 & 0.0 & 41.0 & 7.0 & 1.0 & \textbf{69.0} & 15.0 & 0.0 \\
Task workflows & SR $\uparrow$ & 38.0 & 20.0 & \textbf{22.0} & 42.0 & 0.0 & 0.0 & 51.0 & 3.0 & 9.0 & 32.0 & 0.0 & 0.0 & 38.0 & 25.0 & 14.0 & \textbf{64.0} & \textbf{33.0} & 21.0 \\
Unseen objects & SR $\uparrow$ & 48.0 & 60.0 & 20.0 & 26.0 & 18.0 & 16.0 & 63.0 & 46.0 & 26.0 & \textbf{78.0} & 52.0 & 4.0 & 63.0 & \textbf{65.0} & 19.0 & \textbf{78.0} & 59.0 & \textbf{43.0} \\
\midrule
\rowcolor{gray!15}\multicolumn{20}{l}{\textbf{Long Horizon}} \\
Long horizon & SR $\uparrow$ & 85.0 & 0.0 & 0.0 & 29.0 & 2.0 & 0.0 & 96.0 & 0.0 & 0.0 & 93.0 & 0.0 & 0.0 & 61.0 & \textbf{4.0} & \textbf{1.0} & \textbf{99.0} & 0.0 & 0.0 \\
\midrule
\rowcolor{gray!30}\textbf{Avg} & SR $\uparrow$ & 64.3 & 35.6 & 24.5 & 50.3 & 23.6 & 9.5 & 76.9 & 23.5 & 12.5 & 75.0 & 31.6 & 16.7 & 60.1 & 35.7 & 20.9 & \textbf{82.2} & \textbf{42.7} & \textbf{29.5} \\
\bottomrule
\end{tabular}
\end{adjustbox}
\endgroup
\end{table*}

Across all 3,400 VLA-Arena rollouts, ActionPiece reaches
\vlaLZero\%/\vlaLOne\%/\vlaLTwo\% on L0/L1/L2 and an equal-weight 33-cell mean
of \vlaMean\%, the highest overall average among the compared models.
As on SimplerEnv, this aggregate advantage accompanies variation in task-level
performance. Table~\ref{tab:vla-arena} also reports cumulative safety costs.
Despite training only on L0, ActionPiece leads the L1 and L2 averages by
7.0 and 5.0 points. Leading L2 results on static obstacles, dynamic obstacles,
and unseen objects extend this advantage to obstacle-constrained execution
and object generalization.

\subsection{Component Ablation}
\label{sec:component-ablation}

We evaluate the two physical-order objectives defined in
Sections~\ref{sec:rank-preservation} and~\ref{sec:quantization-regularization}.
Table~\ref{tab:component-ablation} adds each objective individually and both
together to standard RVQ. Standard RVQ reaches
90.9\% on LIBERO and 60.4\% on LIBERO-Plus. Adding physical rank preservation raises
these scores to 93.8\% and 65.4\%; adding only quantization regularization
yields 92.9\% and 62.4\%. Combining both components achieves the highest
overall success on both benchmarks: 94.8\% and 68.8\%.

The same ablations raise PRC from $0.902$ for standard RVQ to $0.947$
with physical rank preservation and $0.916$ with quantization regularization;
the combined model reaches $0.953$. The policy improvements thus accompany better physical neighborhood
preservation. Physical rank preservation contributes the larger individual
gain; adding quantization regularization further raises LIBERO-Plus success
by 3.4 percentage points, supporting their joint use.
Combining the objectives exceeds either alone in six LIBERO-Plus categories,
including a camera-shift gain from 35.5\% with PRP to 45.1\%.
This complementarity supports guiding both feature geometry and discrete
code assignment with physical relationships.

% Author-confirmed component ablation, 2026-09-13.
\begin{table*}[!t]
\centering
\begingroup
\papertablestyle
\setlength{\tabcolsep}{3.2pt}
\caption{\textbf{Component ablation on LIBERO and LIBERO-Plus.}
Starting from standard RVQ, we add physical rank preservation (PRP),
quantization regularization (QR), or both. Success rates are reported in percent. Best results are bold.}
\label{tab:component-ablation}
\resizebox{\textwidth}{!}{%
\begin{tabular}{@{}lccccc@{\hspace{8pt}}cccccccc@{}}
\toprule
& \multicolumn{5}{c}{\textbf{LIBERO}}
& \multicolumn{8}{c}{\textbf{LIBERO-Plus}} \\
\cmidrule(lr){2-6}\cmidrule(lr){7-14}
\textbf{Variant} & Spatial & Object & Goal & Long & Overall
& Camera & Robot & Language & Light & Bkg. & Noise & Layout & Overall \\
\midrule
Standard RVQ
& 91.8 & 98.2 & 91.6 & 82.0 & 90.9
& 31.5 & 43.9 & 75.3 & 86.0 & 88.0 & 42.5 & 72.9 & 60.4 \\
+ PRP
& \textbf{96.2} & 97.4 & 93.0 & 88.4 & 93.8
& 35.5 & 45.6 & \textbf{83.0} & 85.9 & 90.3 & 55.1 & 77.1 & 65.4 \\
+ QR
& 95.4 & 97.2 & 91.2 & 87.8 & 92.9
& 33.0 & 38.8 & 77.7 & 90.8 & 89.7 & 50.2 & 74.3 & 62.4 \\
\rowcolor{gray!30}\textbf{\methodname (both)}
& \textbf{96.2} & \textbf{99.0} & \textbf{93.4} & \textbf{90.6} & \textbf{94.8}
& \textbf{45.1} & \textbf{49.6} & 82.2 & \textbf{91.9} & \textbf{91.4} & \textbf{57.6} & \textbf{78.0} & \textbf{68.8} \\
\bottomrule
\end{tabular}%
}
\endgroup
\end{table*}

\section{Conclusion}

Action tokenization defines the representation that an autoregressive policy learns to predict. We examined how reconstruction errors alter the action variations across demonstrations, and introduced PRC to evaluate physical distance ordering in a common decoded action space alongside reconstruction accuracy. ActionPiece applies this perspective through joint physical rank preservation and quantization regularization. Controlled comparisons and ablations show improvements in both physical structure and downstream success, while four benchmarks cover different action-data sources and generalization settings. These results support incorporating relationships among actions into tokenizer learning for control.

\FloatBarrier

\bibliographystyle{plainnat}
\bibliography{custom}
\clearpage
\appendix
\setcounter{figure}{0}
\renewcommand{\thefigure}{A\arabic{figure}}

\section{Alternative Training Objectives}
\label{app:tokenizer-variants}

We compare alternative objectives added to the standard RVQ tokenizer to
examine how different supervision affects reconstruction, physical structure,
and policy success. Table~\ref{tab:tokenizer-variants} reports measurements for these variants
and the component ablations, using Qwen3-VL-4B at 30K policy training steps.

\paragraph{SIGReg.}
We apply Sketched Isotropic Gaussian Regularization (SIGReg) from
LeJEPA~\citep{balestriero2025lejepa} to encoder features, encouraging their
distribution to match an isotropic Gaussian.

\paragraph{Temporal.}
This objective matches the temporal changes of reconstructed actions to those
of demonstrations. For translation and gripper commands, it compares
first differences between adjacent time steps in the reconstructed and target
sequences. For rotation, it compares the relative quaternion rotations between
adjacent steps. Thus, it supervises the demonstrated pattern of change,
including its magnitude and direction, rather than penalizing motion itself.

\paragraph{Neighborhood.}
This objective attracts physically similar actions in representation space.
For each anchor $i$, we select the ten physically nearest actions in the
minibatch, denoted by $\mathcal N_{10}(i)$. With larger nonnegative weights
$w_{ij}$ for physically closer neighbors, the objective is
\begin{equation}
\mathcal L_{\mathrm{nei}}=
\frac{\sum_i\sum_{j\in\mathcal N_{10}(i)}w_{ij}\,d_{\mathrm{lat}}^2(i,j)}
{\sum_i\sum_{j\in\mathcal N_{10}(i)}w_{ij}},
\label{eq:neighborhood-attraction}
\end{equation}
where $d_{\mathrm{lat}}$ is the distance between action representations.
Neighborhood supervises attraction to nearby actions; physical rank
preservation additionally compares a nearby action with a farther one to
supervise their relative order (Section~\ref{sec:rank-preservation}).

% Confirmed objective variants; values read from the author workbook.
\begin{table*}[!htpb]
\centering
\begingroup
\papertablestyle
\setlength{\tabcolsep}{7pt}
\caption{\textbf{Training-objective variants of standard RVQ.}
Each ``+'' row adds only the named objective to standard RVQ.
PRP and QR denote physical rank preservation and quantization regularization.
Policies use Qwen3-VL-4B with 30K training steps. Success rates are in percent;
MSE is scaled by $10^4$. Best values are bold, determined before rounding.}
\label{tab:tokenizer-variants}
\begin{tabular}{@{}lcccc@{}}
\toprule
Variant & MSE $\downarrow$ & PRC $\uparrow$ & LIBERO Overall $\uparrow$ & LIBERO-Plus Overall $\uparrow$ \\
\midrule
Standard RVQ & 4.06 & 0.902 & 90.9 & 60.4 \\
+ SIGReg & 3.77 & 0.950 & 93.3 & 66.5 \\
+ Neighborhood & 5.09 & 0.889 & 94.3 & 63.5 \\
+ Temporal & 3.83 & 0.924 & 94.8 & 66.1 \\
+ PRP & 3.63 & 0.947 & 93.8 & 65.4 \\
+ QR & 3.92 & 0.916 & 92.9 & 62.4 \\
\rowcolor{gray!30}\textbf{ActionPiece (PRP + QR)} & \textbf{3.32} & \textbf{0.953} & \textbf{94.8} & \textbf{68.8} \\
\bottomrule
\end{tabular}
\endgroup
\end{table*}

These objectives target different properties of the codec. Temporal supervises
changes within a decoded action chunk, Neighborhood attracts representations
across chunks, and physical rank preservation supervises relative distances.
The results illustrate why both reconstruction and neighborhood measurements
are useful: the Neighborhood variant improves policy success over standard RVQ
while its PRC decreases, whereas ActionPiece improves both PRC and success.
ActionPiece achieves the lowest MSE, highest PRC, and highest overall success
on both benchmarks among the variants shown.

\section{Implementation Details}
\label{app:implementation}

\paragraph{ActionPiece configuration.}
We use the same ActionPiece configuration for LIBERO, LIBERO-Plus,
VLA-Arena, and SimplerEnv, as summarized in Table~\ref{tab:actionpiece-settings}.
Other tokenizers use the configurations specified in their original papers
or released codebases.

\begin{table}[!htbp]
\centering
\begingroup
\papertablestyle
\caption{\textbf{ActionPiece configuration and objective weights.}}
\label{tab:actionpiece-settings}
\begin{tabular}{@{}lc@{}}
\toprule
Setting & Value \\
\midrule
Action chunk length $H$ & 8 \\
Output tokens $T$ & 16 \\
Latent slots $S$ & 16 \\
RVQ levels $Q$ & 1 \\
Codebook size $K$ & 512 \\
\midrule
Reconstruction MSE weight & 1 \\
Commitment weight $\lambda_{\mathrm{vq}}$ & 0.25 \\
PRP weight $\lambda_r$ & $6.25\times10^{-4}$ \\
QR weight $\lambda_q$ & $6.25\times10^{-4}$ \\
\bottomrule
\end{tabular}
\endgroup
\end{table}

\subsection{Differentiating Codeword Selection}
\label{app:quantization-estimator}
With $Q=1$, each of the 16 latent slots selects one codeword. Let $u_{i,s}$
be the projected encoder feature and $\Delta_{i,s,c}=\|u_{i,s}-e_c\|_2^2$
its squared distance to codebook entry $c$. Hard assignment selects a code;
to differentiate the representation distance, we use a straight-through
estimator with soft codeword probabilities~\citep{oord2017vqvae,jang2016gumbelsoftmax}.
Let $e_{i,s,\mathrm{hard}}$ be the selected code contribution after the
quantizer output projection, and let $e_{i,s,\mathrm{soft}}$ be the output
projection of the soft codebook average. Here $\operatorname{sg}$ denotes stop-gradient. With
$s_{i,s}=\max\{\operatorname{sg}[\operatorname{Std}_c
(\Delta_{i,s,c})],10^{-6}\}$, we form
\begin{equation}
\begin{aligned}
\pi_{i,s,c}
&=\operatorname{softmax}_c\!\left(
-\frac{\Delta_{i,s,c}}{s_{i,s}\tau_q}\right),\\
\widetilde e_{i,s}
&=e_{i,s,\mathrm{soft}}
+\operatorname{sg}\!\left[
e_{i,s,\mathrm{hard}}-e_{i,s,\mathrm{soft}}
\right].
\end{aligned}
\label{eq:codeword-probabilities}
\end{equation}
We use $\tau_q=1$ and sample no Gumbel noise. Thus $\widetilde e$ has the
selected hard code contribution as its forward value and the soft assignment
for gradient computation. The reconstruction objective uses an identity
straight-through estimator, and the codebook is updated by exponential moving
average (EMA).

\subsection{Multiple Quantization Levels}
\label{app:multiple-levels}
In our experiments, $Q=1$ and $Q=2$ yield similar downstream policy performance.
We use $Q=1$ for the final ActionPiece configuration.
For $Q>1$, we apply the same differentiable codeword selection at each
residual level. Let $q_{i,s}^{(r)}=\sum_{\ell=1}^{r}\widetilde
e_{i,s}^{(\ell)}$ be the cumulative contribution after the corresponding
output projections, and let $\bar q_{i,s}^{(r)}$ be LayerNorm and
unit-normalized. We extend Eq.~\eqref{eq:quantized-geometry} by averaging
over the partial reconstructions at all residual levels:
\begin{equation}
d_{\mathrm{quant}}^2(i,j)
=\frac{1}{Q}\sum_{r=1}^{Q}\frac{1}{S}\sum_{s=1}^{S}
\left\|\bar q_{i,s}^{(r)}-\bar q_{j,s}^{(r)}\right\|_2^2.
\label{eq:prefix-geometry}
\end{equation}
At $Q=1$, this expression reduces to Eq.~\eqref{eq:quantized-geometry}.

\subsection{Physical-Neighbor Selection}
\label{app:neighbor-sampling}

For each anchor in the training batch, we exclude the anchor itself and rank
the remaining action chunks by increasing physical distance. We select the
action at the 5th percentile ($q_{0.05}$) as the positive (near) example and
the action at the 95th percentile ($q_{0.95}$) as the negative (far) example.
These percentiles are computed separately for each anchor over the other
actions in its minibatch. Both physical rank preservation and quantization
regularization use this same pair.
The former compares representation distances $D_{ij}$, while the latter
compares codeword probabilities using Jensen--Shannon divergence. We use
natural logarithms without additional normalization for the divergence.
The discrete action vocabulary and decoding procedure are unchanged by
the two training objectives.

\end{document}